\documentclass[10pt]{article}
\usepackage[preprint]{tmlr}
\usepackage[T1]{fontenc}
\usepackage{microtype}
\usepackage{amsmath,amssymb}
\usepackage{booktabs}
\usepackage{graphicx}
\usepackage{xcolor}
\usepackage{hyperref}
\hypersetup{colorlinks=true,linkcolor=black,citecolor=blue!50!black,urlcolor=blue!50!black}
\usepackage{url}

\newcommand{\vect}[1]{\boldsymbol{#1}}

\title{Gate Initialisation versus Recurrent Dropout and Zoneout in Learning Long Dependencies: A Controlled Study of LSTM and GRU Networks on Synthetic Tasks}

\author{\name Oleksandr Kholodniak \email o.o.kholodniak@khai.edu \\
      \addr National Aerospace University ``Kharkiv Aviation Institute'', Kharkiv, Ukraine
      \AND
      \name Oleksandr Prokhorov \email o.prokhorov@khai.edu \\
      \addr National Aerospace University ``Kharkiv Aviation Institute'', Kharkiv, Ukraine}

\begin{document}

\maketitle

\begin{abstract}
Dropout and zoneout are routinely added to long short-term memory (LSTM) and gated recurrent unit (GRU) networks, but, to
our knowledge, their placements have not been crossed with gate initialisation in a factorial design, and it is not known whether
early gradient propagation tells which runs will learn long dependencies. We trained one-layer LSTMs and GRUs on the
adding problem and on delayed recall in a factorial grid of 720 runs, crossing conventional and chrono gate
initialisation with up to six recurrent regulariser placements at one rate each, with pre-specified follow-ups, and
recorded how sensitive the final hidden state was to the input at every lag. Because both tasks supply fresh or
exhaustive training data, the study measures the regularisers' effects on optimisation and does not assess generalisation from a finite
training set. At lengths
chosen to make conventional initialisation difficult, all 360 chrono-initialised runs reached the validation criterion
within 5000 updates, as did 119 of 120 unregularised runs with a horizon about four times too short or too long,
whereas conventionally initialised networks learned slowly or did not reach the criterion within the budget at the
longer lengths. Under conventional initialisation, dropout generally delayed learning and zoneout helped only in some
conditions. The early input sensitivity added little to predicting which conditions learn once the early validation
loss was modelled per task, architecture and initialisation. Within conditions, a flatter early decay went with the
seeds that learned, but in a pre-specified test the association was modest (concordance 0.61), condition-dependent and
not robust to resampling seeds. The results support chrono initialisation, even with a rough horizon, on such
tasks.
\end{abstract}

\section{Introduction}

In a recurrent network trained by backpropagation through time (BPTT), the computations that process an input seen $T$
steps before the loss receive direct credit only through the gradient carried back over those steps \citep{werbos1990}. When the Jacobians along this
path contract, the gradient decays exponentially with the lag and long dependencies become hard to learn
\citep{bengio1994,pascanu2013}. Gated architectures, the long short-term memory (LSTM) \citep{hochreiter1997,gers2000} and the gated
recurrent unit (GRU) \citep{cho2014}, mitigate this with multiplicative gates whose biases set how long a unit keeps its state, and the
initial value of these biases matters: a positive forget-gate bias is standard for LSTMs
\citep{gers2000,jozefowicz2015}, and chrono initialisation draws the gate biases so that the units' memory timescales
cover the horizon of the task \citep{tallec2018}. \citet{tallec2018} showed that chrono
initialisation improves the learning of long dependencies by LSTMs over a forget-gate bias of one, with faster
convergence on the adding problem and a clearer advantage on copy tasks.

The same networks are usually regularised with stochastic noise in the recurrence: dropout on the recurrent inputs of
the gates, with a new mask at every step or with one mask per sequence \citep{gal2016}, dropout of the cell state with
one mask per sequence \citep{moon2015}, dropout on the candidate update \citep{semeniuta2016}, or zoneout, which keeps a random subset of units at their previous value \citep{krueger2017}.
These placements differ in how they interact with the gradient path. Zoneout was introduced partly because identity
connections should help the gradient propagate, and \citet{krueger2017} showed, in one figure for one task, that after
one epoch the gradient of an LSTM with zoneout reached earlier steps than with dropout of the cell state or without
regularisation. What has not been
measured is how large these effects are compared with gate initialisation, whether they hold for LSTMs and GRUs alike,
and whether the reach of the gradient, measured early in training, tells which runs will eventually learn.

This last question matters in practice because training recurrent networks on long dependencies is slow and its outcome
varies between seeds. If early gradient reach predicted success, it could be used to stop unpromising runs or to choose
between configurations before training has gone far. Comparable signals are already used for this: early validation
performance for fine-tuning runs of pretrained language models \citep{dodge2020}, extrapolated learning curves and
successive halving for hyperparameter search \citep{domhan2015,li2018hyperband}, and zero-cost proxies computed from
gradients at initialisation for architectures \citep{abdelfattah2021}. Theory already links propagation timescales at
initialisation to trainability \citep{chen2018,gilboa2019}, and the decay of gradient norms has been estimated during
training for other purposes \citep{aicher2019,chernodub2016}. Whether such a measurement adds information beyond what
the training curve itself shows is an empirical question.

\begin{figure}[tbp]
  \centering
  \includegraphics[width=\textwidth]{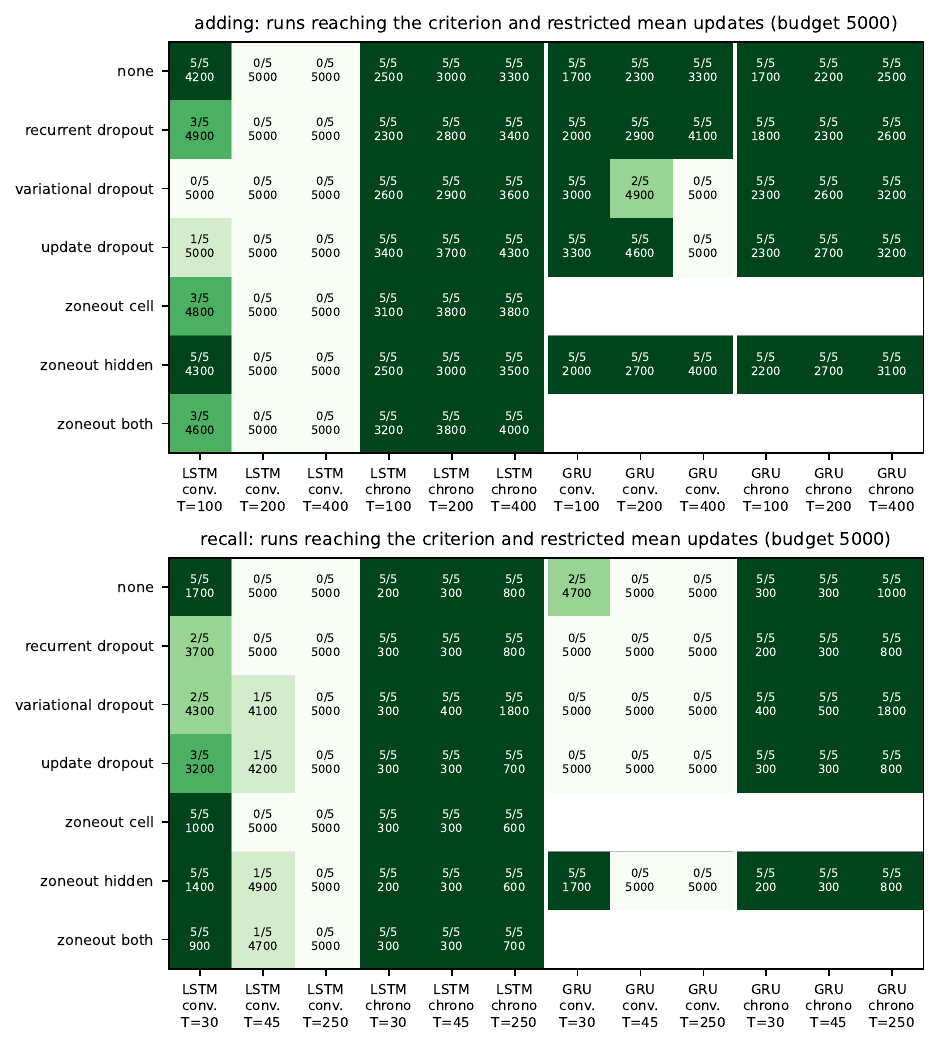}
  \caption{Chrono initialisation reached the criterion in every main-grid run; conventional initialisation often did not
    at the longer lengths, where the regulariser could change the outcome. Each cell gives the runs that reached the
    validation criterion within 5000 updates (of five seeds) and the restricted mean number of updates to the criterion
    (runs that did not reach it count as 5000); darker shading means more runs reached it. Blank cells: cell-state and
    both-state zoneout, not defined for the GRU. Columns combine architecture, initialisation (conv.\ = conventional)
    and length.}
  \label{fig:outcomes}
\end{figure}

We answer these questions with a factorial grid of 720 training runs of one-layer LSTM and GRU networks on two
long-dependency tasks, crossing gate initialisation with six regulariser placements for the LSTM and four for the GRU,
at one rate each, and five seeds, followed by runs with more seeds, a longer budget and mismatched chrono horizons
(Table~\ref{tab:design}). Both tasks supply fresh or exhaustive training data, so the study measures how the
regularisers affect optimisation; their protection against overfitting to a finite training set is outside its scope. These networks
are trained with a library whose BPTT passes finite-difference checks and agrees with PyTorch to about $10^{-16}$ in
double precision \citep{rustlstm}; 120 further runs repeat the unregularised conditions with PyTorch's built-in cells. The contributions are:
\begin{enumerate}
  \item We find that gate initialisation produced a larger contrast in whether and how fast the networks learned than
    any regulariser at the rates tested, although the regularisers still changed attainment and learning time in some
    conditions (Figure~\ref{fig:outcomes}). For the LSTM on the adding problem this replicates \citet{tallec2018};
    it extends their comparison to the GRU, to delayed recall, to the crossing with regularisers and to mismatched
    horizons, under which chrono initialisation kept its advantage with a horizon about four times too short or too long.
  \item We measure how each regulariser placement changes the slope of the input sensitivity with lag after 100 updates
    (Table~\ref{tab:effects}), and find that a flattening by zoneout of similar size is already present in evaluation
    mode before training.
  \item We find that this slope adds little to predicting which \emph{conditions} learn once the early validation loss is
    modelled per architecture, initialisation and task (Table~\ref{tab:prediction}). Within a condition it is
    associated with which \emph{seeds} learn, but a test specified before those runs supported this only when
    resampling conditions, not seeds (Figure~\ref{fig:within}).
  \item We release the code, the per-run results and the gradient curves of every run.
\end{enumerate}

Figure~\ref{fig:outcomes} summarises the main grid.

\section{Related work}

\paragraph{Regularising the recurrence.}
\citet{zaremba2014} applied dropout only to non-recurrent connections because dropping the recurrent state at every
step erased memory. \citet{moon2015} kept one mask per sequence and applied it to the cell state, and \citet{gal2016} derived the same per-sequence
(variational) mask from approximate Bayesian inference and applied it to the input and to the recurrent state where they
enter the gates.
\citet{semeniuta2016} argued that dropout should act on the candidate update rather than on the state, so that what is
already stored is never erased. Zoneout replaces dropped units by their previous values instead of zeros
\citep{krueger2017}. These works compare placements by task performance; only \citet{krueger2017} report a gradient
measurement, a normalised snapshot of cell-state gradient norms on one task.

\paragraph{Gating and initialisation.}
A forget-gate bias of one is a standard default for LSTMs \citep{gers2000,jozefowicz2015}, and large studies of LSTM
variants found the forget gate to be among their most important components \citep{greff2017}. Chrono initialisation
links the gate biases to the timescales a task requires \citep{tallec2018}. \citet{gu2020} found chrono initialisation
sensitive to the choice of $T_{\max}$ and proposed a uniform gate initialisation that needs no horizon, together with a
refine gate against saturation, and \citet{chandar2019} proposed non-saturating recurrent units. Other approaches to long memory constrain or initialise the
recurrent matrix itself \citep{le2015,arjovsky2016,henaff2016}, and recent recurrences designed for long sequences set
the timescales of the state through optimal polynomial projections \citep{gu2020hippo} or through linear recurrences
with eigenvalues initialised near the unit circle and then learned \citep{orvieto2023}. Our study concerns the classic
non-linear gated cells, in which gate biases play this role.

\paragraph{Gradient propagation and trainability.}
Exponential decay of gradients in recurrent networks was analysed by \citet{bengio1994} and \citet{pascanu2013}.
Mean-field theory predicts propagation timescales from initialisation and relates them to which networks can be trained
\citep{chen2018,gilboa2019}. \citet{zucchet2024} show that long memory makes optimisation difficult even when gradients
neither vanish nor explode, and h-detach modifies the gradient flowing through the LSTM hidden path to aid the cell path
\citep{arpit2019}. The decay rate of gradient norms has been estimated during training to choose truncation lengths
\citep{aicher2019}, and per-example estimates of the long-term part of the gradient have been used to assemble
mini-batches \citep{chernodub2016}. We use such measurements not to change training but as an observable, and test whether they
predict its outcome.

\section{Methods}

\subsection{Networks}\label{sec:networks}

All networks have one recurrent layer of $H = 64$ units and a linear output layer applied to the last hidden state
$\vect{h}_T$. The LSTM computes, for input $\vect{x}_t$,
\begin{equation}
  \vect{c}_t = \vect{f}_t \odot \vect{c}_{t-1} + \vect{i}_t \odot \vect{g}_t, \qquad
  \vect{h}_t = \vect{o}_t \odot \tanh(\vect{c}_t),
\end{equation}
where the input, forget and output gates $\vect{i}_t, \vect{f}_t, \vect{o}_t$ are logistic and the candidate
$\vect{g}_t$ is a $\tanh$ of affine functions of $\vect{x}_t$ and $\vect{h}_{t-1}$, and $\odot$ is the elementwise
product. We write the GRU in the convention of \citet{chung2014}, with the reset gate applied before the recurrent matrix
as in \citet{cho2014}:
\begin{equation}
  \vect{n}_t = \tanh\bigl(W_n \vect{x}_t + U_n(\vect{r}_t \odot \vect{h}_{t-1}) + \vect{b}_n\bigr), \qquad
  \vect{h}_t = (1 - \vect{z}_t) \odot \vect{h}_{t-1} + \vect{z}_t \odot \vect{n}_t,
\end{equation}
with logistic reset and update gates $\vect{r}_t$ and $\vect{z}_t$. Weight matrices are drawn from $U(-0.1, 0.1)$,
output weights from $U(-a, a)$ with $a = \sqrt{2/(H + O)}$ for $O$ outputs, and biases are zero except as stated next.
The networks were implemented and trained with rust-lstm 0.10.0 \citep{rustlstm}, whose test suite checks every
backward pass, including those of dropout and zoneout in training mode, against central finite differences, and in which
LSTM and GRU outputs and gradients agree with PyTorch to about $10^{-16}$ in double precision on reference cases.
The initialisation contrast was also rerun with PyTorch's own modules (Section~\ref{sec:pytorch}).

\paragraph{Gate initialisation.}
\emph{Conventional} initialisation sets the LSTM forget-gate bias to 1 and all other biases to 0; the GRU update gate
then starts at $\sigma(0) = 0.5$, so, unlike the LSTM, the conventional GRU has no bias towards keeping its state. \emph{Chrono} initialisation \citep{tallec2018} draws a timescale
$u_j \sim U[1, T_{\max} - 1]$ for every unit $j$ and sets the LSTM biases to $b_{f,j} = \log u_j$ and
$b_{i,j} = -\log u_j$. For the GRU equation above, the corresponding choice is $b_{z,j} = -\log u_j$, so that
$z_j \approx 1/(1 + u_j)$ and unit $j$ retains about a fraction $1 - 1/(1+u_j)$ of its state per step. In the main grid
$T_{\max}$ equals the sequence length $T$; a follow-up uses $T_{\max} = \max(3, \lfloor T/4 \rfloor)$ and $4T$.

\paragraph{Regularisers.}
Six placements for the LSTM and four for the GRU are compared with an unregularised baseline:
\begin{itemize}
  \item \emph{recurrent dropout}: dropout on the recurrent input $\vect{h}_{t-1}$ of the gates with a new mask at every
    step;
  \item \emph{variational dropout}: the same with one mask per sequence \citep{gal2016}; Keras's
    \texttt{recurrent\_dropout} is of this kind;
  \item \emph{update dropout}: dropout on the candidate $\vect{g}_t$ of the LSTM or $\vect{n}_t$ of the GRU
    \citep{semeniuta2016};
  \item \emph{zoneout} \citep{krueger2017} of the LSTM cell state, of the hidden state, or of both (GRU: hidden state
    only), which replaces a new state $\tilde{\vect{s}}_t$ by $\vect{d}_t \odot \vect{s}_{t-1} + (1 - \vect{d}_t) \odot
    \tilde{\vect{s}}_t$ with Bernoulli masks $\vect{d}_t$.
\end{itemize} Dropout masks are inverted (scaled by
$1/(1-p)$) and never touch the carried state. Rates are $p = 0.25$ for dropout and $q = 0.15$ for zoneout; at
evaluation, dropout is off and zoneout uses its expectation $\vect{d}_t = q$, which adds a fixed leak to the carried
state.

\subsection{Tasks}

In the \emph{adding problem} \citep{hochreiter1997}, in the form of \citet{le2015}, each input is a pair $(v_t, m_t)$ with $v_t \sim U[0, 1]$ and a marker
$m_t$ that equals one at one position in each half of the sequence; the target is the sum of the two marked values,
predicted at the last step and scored by squared error. Predicting the mean, 1, gives a mean squared error of $1/6$.
In \emph{delayed recall} one of eight symbols is presented as a one-hot vector at the first step, the inputs are zero for
the next $T - 2$ steps, and a query flag at the last step asks for the symbol, scored by softmax cross-entropy. For a
given length recall has only eight distinct sequences. Because adding draws fresh sequences for every batch and recall
validates on all of its eight sequences, neither task assesses the usual effect of regularisation on generalisation
from a finite training set.

\subsection{Training and endpoint}

Networks are trained with Adam \citep{kingma2015} at learning rate $10^{-3}$, batches of 32 fresh sequences, and the
recurrent and output gradients clipped jointly to norm 1. Every 100 updates the network is evaluated on fixed validation
data: 1024 adding sequences, or all eight recall sequences. A run \emph{reaches the criterion} at the first evaluation at
which the validation mean squared error falls below 0.01 (a 94\,\% reduction from the constant prediction) or all eight
symbols are recalled, and training then stops; otherwise it stops at the budget of 5000 updates (20\,000 in one
follow-up). Reaching the criterion is the endpoint throughout. In the main grid a final score on 4096 fresh adding
sequences confirms it for 243 of the 262 adding runs that reached it (169 of the 180 chrono-initialised ones); the
remaining 19 have test errors just above 0.01.

\subsection{Gradient probes}

At updates 0, 100, 250, 500, 1000 and 2000, unless the run has already stopped, the gradients of a fixed probe batch of
64 sequences are recorded at every lag $k = 0, \dots, T - 1$, where lag $k$ denotes the input $k$ steps before the last
one. The main quantity is the input sensitivity of projections of the final hidden state,
\begin{equation}
  G(k) = \frac{1}{4} \sum_{j=1}^{4} \frac{1}{64} \sum_{b=1}^{64}
    \log \left\| \frac{\partial\, \vect{u}_j^{\top} \vect{h}_{T,b}}{\partial \vect{x}_{T-k,b}} \right\|,
\end{equation}
where $b$ indexes the sequences of the probe batch and $\vect{u}_1, \dots, \vect{u}_4$ are fixed random unit vectors.
$G(k)$ is independent of the loss and of the output layer. Its trend with lag is summarised by the slope $s$ of a
least-squares line of $G(k)$ on $k$ over $1 \le k \le T/2$, in natural-log units per step. This window was fixed in the
first analysis code; it excludes the lag of the recall symbol and all but one possible lag of the first adding marker,
so $s$ describes the trend over the more recent half of the sequence, and slopes over $1 \le k \le T/4$ and
$1 \le k \le T - 1$ are computed as sensitivity analyses. We use $s$ as a descriptive summary of input sensitivity: a value near zero means little relative
attenuation over the window and says nothing about the earliest inputs, which lie outside it. Only for a negative slope and a
curve close to exponential is $-1/s$ the lag over which the sensitivity shrinks by a factor of $e$. We also record the gradient of each sequence's own
loss and, as task-specific checks, the derivative of the adding prediction with respect to the two marked values and
of the true-class logit with respect to the recall symbol. Probes are taken in evaluation mode and in training mode, the
latter averaged over eight draws of the dropout and zoneout masks from a random stream separate from training. In
evaluation mode dropout is off, so its probes show only the dynamics it induced during training, whereas the expected
zoneout mask enters the probe directly.

\subsection{Experimental design}

Table~\ref{tab:design} summarises the plans. Two calibration plans with unregularised networks and seeds 0 and 1
selected the lengths of the main grid so that each task has a length that most conditions solve, one in transition, and
one that most fail without chrono initialisation. The main grid crosses architecture, initialisation, regulariser,
task and length with five seeds; of its 144 conditions, 53, all conventionally initialised, had fewer than five of five
runs reaching the criterion. Three follow-ups were specified, with their analyses, before they were run (repository
commit \texttt{a54e1cb}): more seeds for
the 13 conditions in which some but not all seeds reached the criterion, a fourfold budget for the 32 of the 53 at the
two shorter lengths of each task, and chrono initialisation with mismatched horizons. The 48 unregularised calibration
runs with seeds 0 and 1 at the main-grid lengths coincide with main-grid runs and gave identical results. The long-budget
runs repeat main-grid runs for their first 5000 updates and continue them to 20\,000; the 25 that reached the criterion
within 5000 updates are identical to their main-grid counterparts. All random streams are seeded for each fixed run configuration
and implementation. Runs used rust-lstm 0.10.0 and rustc 1.86 on one Apple M1 Pro.

\begin{table}[tbp]
  \centering
  \caption{Experimental plans. Conditions combine task, length, architecture, initialisation and regulariser; the
    GRU has five regulariser settings (no cell-state zoneout), the LSTM seven.}
  \label{tab:design}
  \small
  \begin{tabular}{@{}lp{8.3cm}lrr@{}}
    \toprule
    Plan & Content & Seeds & Runs & Budget \\
    \midrule
    Calibration & unregularised; adding $T \in \{50,\dots,400\}$, recall $T \in \{10,\dots,500\}$ & 0--1 & 104 & 5000 \\
    Main grid & 2 architectures $\times$ 2 initialisations $\times$ 5 or 7 regularisers $\times$ 6 lengths & 0--4 & 720 & 5000 \\
    More seeds & the 13 main-grid conditions with mixed outcomes & 5--19 & 195 & 5000 \\
    Long budget & 32 conventional conditions below 5/5 at shorter lengths & 0--4 & 160 & 20\,000 \\
    Horizon mismatch & unregularised, chrono with $T_{\max} \approx T/4$ or $4T$ & 0--4 & 120 & 5000 \\
    PyTorch replication & the 24 unregularised main-grid conditions, PyTorch \texttt{nn.LSTM}/\texttt{nn.GRU}; outcomes only & 0--4 & 120 & 5000 \\
    \bottomrule
  \end{tabular}
\end{table}

Main-grid lengths are $T \in \{100, 200, 400\}$ for adding and $T \in \{30, 45, 250\}$ for recall.

\subsection{Statistical analysis}

\paragraph{Landmark prediction across conditions.}
Following the landmark approach to dynamic prediction \citep{vanhouwelingen2007}, prediction is made from a fixed point
in training, the landmark at update 100: every main-grid run not yet at the criterion there (all 720) is followed to the
end of its budget. A gradient-free baseline predicts reaching the criterion from
task, length, architecture, initialisation, regulariser and the log validation loss at update 100; the full model adds
the slope $s$ at update 100 in evaluation mode. The \emph{additive} baseline enters all factors as main effects. It
was committed with the first analysis code while the main grid was running.\footnote{Commit \texttt{f9cf778}, three
minutes after the main grid started.} A \emph{stratified} baseline, added after a review of our first analysis, gives each
task $\times$ architecture $\times$ initialisation stratum its own intercept and its own effects of log length and log
landmark loss. Several conditions separate the outcomes completely, so unpenalised maximum-likelihood estimates would
not exist \citep{heinze2002}; the models are therefore $L_2$-penalised logistic regressions on standardised features with
inverse penalty strength $C = 1$, which keeps coefficients finite. Predictions are made out of sample, holding out all runs of a condition together in
ten grouped folds (primary) or holding out whole seeds, regulariser families or task lengths \citep{roberts2017}. We
report the reduction in log loss and its 95\,\% interval from resampling conditions, and for the primary comparison
also from resampling conditions and seeds crossed; these intervals hold the out-of-sample predictions fixed.

\paragraph{Ranking within conditions.}
Within each condition with both outcomes, every pair of a run that reached the criterion and one that did not is
scored 1 if the run that reached it has the larger predictor, $\tfrac12$ for ties and 0 otherwise; the mean over pairs is
a concordance (an area under the curve within conditions). For the loss the predictor is $-\log$ of the validation loss
at the landmark, so that a lower loss counts as predicting success. The exploratory analysis used the main grid. The
confirmatory test, written into the project README before the follow-up runs (commit \texttt{a54e1cb}), used only
seeds 5--19, predicted a
concordance above 0.5 for the slope and not for the landmark loss, and specified a 95\,\% interval from resampling
conditions. Because a seed fixes the initial weights and the training data for every condition of a given architecture,
task and length, runs with the same seed are not independent across conditions; we therefore also report intervals that
resample seeds (a pair counts once per draw of each of its two seeds) and that resample conditions and seeds crossed.

\paragraph{Time to the criterion.}
A discrete-time hazard model \citep{singer1993} uses one record per run and evaluation interval from update 200 until
the criterion is reached or the budget ends, and models the probability of reaching the criterion in an interval by a
penalised logistic regression on log time, the additive baseline covariates and the slope, standardised across runs.

\section{Results}

\subsection{In this grid, initialisation decided whether and how fast the tasks were learned}

Figure~\ref{fig:outcomes} shows the main grid. With chrono initialisation all 360 runs in 72 conditions reached the
criterion within 5000 updates; with conventional initialisation 120 of 360 did, and 53 of its 72 conditions had fewer
than five of five. Without regularisation, the
conventionally initialised LSTM reached it in none of ten adding runs at $T = 200$ and 400 and in none of ten recall
runs at $T = 45$ and 250, and the conventionally initialised GRU in 2 of 5 recall runs at $T = 30$ and none at the two
longer recall lengths.

A fourfold budget shows that many of these failures are slow learning rather than inability
(Table~\ref{tab:long}). Of the 160 runs in the 32 conventional conditions below 5/5 at the two shorter lengths of each
task, 25 reached the criterion within 5000 updates and 121 within 20\,000. All 35 LSTM adding runs at $T = 200$ reached it, with medians between 8100 updates
without regularisation and 11\,200 with variational dropout. Recall at $T = 45$ remained hard: without regularisation
3 of 5 LSTM and 5 of 5 GRU runs reached it.

\begin{table}[tbp]
  \centering
  \caption{Long budget: conventionally initialised conditions below 5/5 at 5000 updates, rerun with 20\,000. Runs (of five)
    that reached the criterion by 5000 $\to$ by 20\,000 updates, and median updates among those that reached it.
    Selected conditions; all 32 are tabulated in the repository. rec.\ = recurrent, var.\ = variational, upd.\ = update
    dropout; zo.\ = zoneout of the cell, hidden or both states.}
  \label{tab:long}
  \footnotesize
  \begin{tabular}{@{}llrrrrrrr@{}}
    \toprule
    & & \multicolumn{1}{c}{none} & \multicolumn{1}{c}{rec.} & \multicolumn{1}{c}{var.} & \multicolumn{1}{c}{upd.}
      & \multicolumn{1}{c}{zo.\ cell} & \multicolumn{1}{c}{zo.\ hid.} & \multicolumn{1}{c}{zo.\ both} \\
    \midrule
    LSTM adding $T=200$ & by 5000 $\to$ 20\,000 & 0 $\to$ 5 & 0 $\to$ 5 & 0 $\to$ 5 & 0 $\to$ 5 & 0 $\to$ 5 & 0 $\to$ 5 & 0 $\to$ 5 \\
                        & median            & 8100 & 9000 & 11\,200 & 10\,100 & 8500 & 8700 & 9100 \\
    GRU recall $T=30$   & by 5000 $\to$ 20\,000 & 2 $\to$ 5 & 0 $\to$ 5 & 0 $\to$ 5 & 0 $\to$ 2 & -- & 5 $\to$ 5$^{*}$ & -- \\
    GRU recall $T=45$   & by 5000 $\to$ 20\,000 & 0 $\to$ 5 & 0 $\to$ 0 & 0 $\to$ 5 & 0 $\to$ 0 & -- & 0 $\to$ 2 & -- \\
    LSTM recall $T=45$  & by 5000 $\to$ 20\,000 & 0 $\to$ 3 & 0 $\to$ 1 & 1 $\to$ 1 & 1 $\to$ 1 & 0 $\to$ 2 & 1 $\to$ 3 & 1 $\to$ 3 \\
    \bottomrule
    \multicolumn{9}{@{}l}{\footnotesize $^{*}$Already 5 of 5 within 5000 updates in the main grid; not rerun.}
  \end{tabular}
\end{table}

Chrono initialisation does not owe its advantage to knowing the task length exactly. With
$T_{\max} = \max(3, \lfloor T/4 \rfloor)$ or $4T$, 119 of 120 unregularised runs in 24 conditions reached the criterion
within 5000 updates; the exception was one LSTM recall run at $T = 250$ with the short horizon. On recall the mismatched horizons were somewhat slower than the matched one at short
lengths (for example 420 against 240 updates for the LSTM at $T = 30$ with $T/4$), but at $T = 30$ and 45 their
restricted mean numbers of updates were 4 to 17 times lower than with conventional initialisation (at $T = 30$: 1700
updates for the LSTM and
4680 for the GRU, counting runs that did not reach the criterion as 5000, against 260--540 with mismatched horizons). On
adding the short horizon was the fastest GRU setting at every length. With $T_{\max} = T/4$ and gates held at their
initial biases, the slowest possible unit would retain $(1 - 1/T_{\max})^{T}$, about 1--2\,\% of its state over a whole
sequence at these lengths. The experiment shows that chrono initialisation tolerates this mismatch on tasks with one
long lag.

\paragraph{Replication in PyTorch.}\label{sec:pytorch}
After the main analysis, the 24 unregularised conditions of the main grid (two architectures, two initialisations, six
task lengths, seeds 0--4) were rerun with the LSTM and GRU modules of PyTorch 2.9.1 \citep{paszke2019}, with PyTorch's
default weight initialisation, single precision and independently drawn training data, but the same gate biases, tasks,
optimiser, batch size, clipping, budget, validation schedule and criteria. PyTorch's GRU differs from the cell of
Section~\ref{sec:networks} in two ways: it keeps the old state with weight $z$, so its chrono biases are
$b_{z,j} = \log u_j$, and it applies the reset gate after the recurrent matrix, $\vect{r}_t \odot (U\vect{h}_{t-1} +
\vect{b})$, rather than before it. All 60 chrono-initialised runs reached the criterion, as in the main grid, and 27 of
60 conventionally initialised runs did in each implementation. The number of seeds that reached the criterion was the
same in 21 of the 24 conditions and differed by at most two (LSTM recall at $T = 30$: 5 against 3 of 5); across the 18
conditions in which runs reached the criterion in both implementations, the median numbers of updates had a Spearman
correlation of 0.97. The advantage of chrono initialisation in unregularised learning outcomes thus also appeared with
PyTorch's built-in cells, which is evidence against an artefact of the implementation used for the main grid. Only these
outcomes were replicated: the gradient measurements were not repeated, and the regulariser comparisons rest on the main
implementation alone, because PyTorch's recurrent modules do not provide the recurrent dropout and zoneout placements.

\subsection{Regularisers changed learning speed, and conventional initialisation exposed the difference}

Under conventional initialisation dropout in every placement generally delayed learning, and in some conditions it
prevented it within the budget (Figure~\ref{fig:outcomes}, Table~\ref{tab:long}). The conventional GRU reached the criterion on
adding at $T = 400$ in 5 of 5 runs without regularisation but in none with variational or update dropout. On recall at
$T = 30$ it reached it in 2 of 5 runs without regularisation and in none with any dropout placement within 5000
updates; at $T = 45$ it reached it in 5 of 5 runs within 20\,000 updates without regularisation but in none with
recurrent or update dropout. Zoneout of the hidden state let the same GRU reach recall at $T = 30$ in 5 of 5 runs, and
cell or both-state zoneout shortened the conventional LSTM's restricted mean time on recall at $T = 30$ from 1700 to 960
and 940 updates. Zoneout was not uniformly helpful, however: it slowed the conventional GRU on adding at $T = 400$ (4000
against 3260 updates), lowered the LSTM on adding at $T = 100$ from 5 to 3 of 5 runs within 5000 updates in the cell and
both-state variants, and with 20\,000 updates the GRU reached recall at $T = 45$ in only 2 of 5 runs with hidden zoneout
against 5 of 5 without regularisation. Dropout was not uniformly harmful either: on recall at $T = 45$ the conventional
LSTM reached the criterion within 5000 updates in one of five runs with variational or update dropout and in none
without regularisation.

Under chrono initialisation every regulariser reached the criterion in every run, but learning times still differed:
variational dropout roughly doubled the time on recall at $T = 250$ (1800 against 820 updates for the LSTM, 1760 against
960 for the GRU), whereas update dropout shortened it slightly (740 and 760), as did recurrent dropout for the GRU
(820). These are contrasts at one rate per
regulariser family and do not rank the families in general.

\subsection{Initialisation set the decay of input sensitivity more than the regularisers did}

\begin{figure}[tbp]
  \centering
  \includegraphics[width=\textwidth]{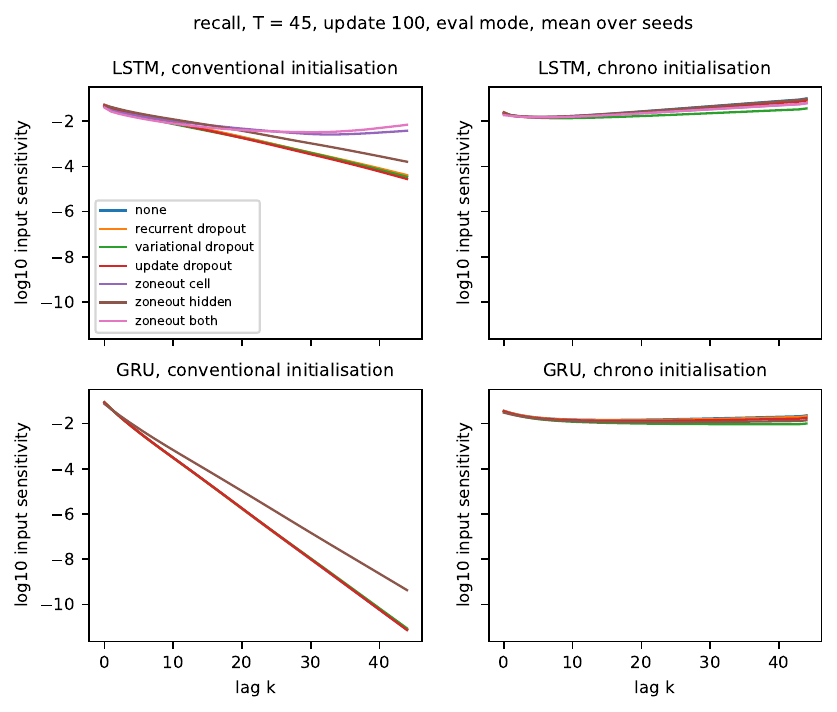}
  \caption{Under conventional initialisation the input sensitivity decays roughly exponentially with lag; chrono
    networks attenuate far less. Input sensitivity of the final hidden state, $G(k)/\ln 10$, against lag $k$ for recall
    at $T = 45$ after 100 updates, evaluation mode, mean over five seeds.}
  \label{fig:curves}
\end{figure}

Without regularisation and before training, the slope of $G(k)$ per step, averaged over the six task lengths, was
$-0.514$ for the conventional GRU, $-0.185$ for the conventional LSTM, $-0.044$ for the chrono GRU and $-0.042$ for the
chrono LSTM. The GRU value is of the order of $\log 0.5 = -0.69$ expected from an update gate at 0.5 in the direct carry
alone. Conventional curves were close to exponential (median $R^2$ of the line at least 0.95 in every task, length and
architecture at update 100), whereas chrono curves were not (median $R^2$ between 0.61 and 0.94) and in short recall
were sometimes increasing; for chrono networks the slope is a descriptive summary rather than a decay rate.
Figure~\ref{fig:curves} shows the resulting contrast for recall at $T = 45$: after 100 updates, the sensitivity of the
unregularised conventional GRU to the first input is about ten orders of magnitude below its sensitivity to the last one
(about eight with hidden zoneout), that of the unregularised conventional LSTM about three (about one with cell or
both-state zoneout), and chrono networks are nearly flat. The task-specific derivatives, averaged over the three lengths
and five seeds of each task in evaluation mode after 100 updates, agree: the derivative of the unregularised adding
prediction with respect to the first marked value was $1.6 \times 10^{-8}$ for the conventional GRU and
$5.7 \times 10^{-6}$ for the conventional LSTM, against $2.8 \times 10^{-3}$ and $5.3 \times 10^{-3}$ with chrono
initialisation, and the derivative of the recall logit with respect to the symbol was $2 \times 10^{-9}$ for the
conventional GRU against 0.07 for the conventional LSTM and 0.23 and 0.41 for the chrono networks. These derivatives show
that early inputs reach the output, not that the networks already use them selectively: for the chrono LSTM the mean
absolute derivative with respect to unmarked values ($5.3 \times 10^{-3}$) was as large as that with respect to the marked
ones, against target derivatives of 1 and 0.

\begin{table}[tbp]
  \centering
  \caption{Change of the slope $s$ at update 100 (evaluation mode) against the unregularised network with the same seed,
    task and length, averaged over lengths within each seed; mean and pointwise 95\,\% $t$ interval over five seeds
    (exploratory). Positive values mean slower decay; rec., var., upd.\ = recurrent, variational, update. Zoneout values
    include the direct effect of its expected mask in evaluation mode, present before training (see text).}
  \label{tab:effects}
  \scriptsize
  \setlength{\tabcolsep}{3pt}
  \begin{tabular}{@{}lrrrr@{}}
    \toprule
    & \multicolumn{2}{c}{Conventional} & \multicolumn{2}{c}{Chrono} \\
    \cmidrule(lr){2-3}\cmidrule(l){4-5}
    Regulariser & LSTM & GRU & LSTM & GRU \\
    \midrule
    rec.\ dropout       & $-0.006$ [$-0.023$, $0.011$] & $-0.017$ [$-0.022$, $-0.012$] & $-0.001$ [$-0.002$, $0.001$] & $-0.001$ [$-0.002$, $0.000$] \\
    var.\ dropout       & $-0.038$ [$-0.057$, $-0.019$] & $-0.060$ [$-0.068$, $-0.052$] & $-0.006$ [$-0.012$, $-0.001$] & $-0.007$ [$-0.013$, $-0.002$] \\
    upd.\ dropout       & $-0.008$ [$-0.032$, $0.016$] & $-0.034$ [$-0.043$, $-0.024$] & $-0.001$ [$-0.004$, $0.001$] & $-0.002$ [$-0.007$, $0.002$] \\
    zoneout, cell       & $0.033$ [$0.015$, $0.050$] & -- & $0.000$ [$-0.006$, $0.006$] & -- \\
    zoneout, hidden     & $0.003$ [$-0.009$, $0.016$] & $0.066$ [$0.062$, $0.070$] & $0.000$ [$-0.001$, $0.001$] & $-0.003$ [$-0.011$, $0.004$] \\
    zoneout, both       & $0.045$ [$0.032$, $0.057$] & -- & $-0.001$ [$-0.006$, $0.005$] & -- \\
    \bottomrule
  \end{tabular}
\end{table}

Table~\ref{tab:effects} gives the change of the slope caused by each regulariser after 100 updates. Under conventional
initialisation variational dropout steepened the decay most for both architectures, update and recurrent dropout less,
and zoneout flattened it where it acts on the path that carries memory: the cell state of the LSTM and the hidden state
of the GRU. A flattening of similar size is present before any training: at update 0, when the networks differ only in
the evaluation-mode zoneout mask, the slope changed by $+0.092$ for the conventional GRU with hidden zoneout and by $+0.028$
and $+0.030$ for the conventional LSTM with cell and both-state zoneout. After training, the contrast combines this
effect of the expected mask with learned dynamics, and the two cannot be separated here. Under chrono initialisation the mean changes were at most about 0.007 in absolute value, the
largest for variational dropout. Measured in training mode before any training, the picture differs: variational dropout
then flattened the curve ($+0.027$ for the GRU and $+0.020$ for the LSTM under conventional initialisation), although it
delayed learning in several of the conditions above. Input sensitivity at initialisation therefore does not rank the
dropout placements by how well they train.

\subsection{The early slope added little to predicting which conditions learned}\label{sec:prediction}

\begin{table}[tbp]
  \centering
  \caption{Landmark prediction of reaching the criterion from update 100 (720 runs, 480 reached it; seeds 2--4 only: 432
    runs, 288 reached it). Log loss of the gradient-free baseline and of the baseline plus the slope $s$, out of sample,
    and the reduction with a 95\,\% interval from resampling conditions; for held-out conditions also from resampling
    conditions and seeds crossed. Intervals hold the out-of-sample predictions fixed; reductions are computed before
    rounding.}
  \label{tab:prediction}
  \small
  \begin{tabular}{@{}llrrrr@{}}
    \toprule
    Baseline & Held out & Baseline & $+\,s$ & Reduction & Interval \\
    \midrule
    additive & conditions (10 folds) & 0.239 & 0.165 & 0.074 & [0.044, 0.104] \\
             & \quad crossed interval & & & & [0.022, 0.120] \\
    stratified & conditions (10 folds) & 0.162 & 0.154 & 0.008 & [$-0.002$, 0.018] \\
             & \quad crossed interval & & & & [$-0.009$, 0.025] \\
    stratified & seeds & 0.141 & 0.130 & 0.011 & [0.002, 0.022] \\
    stratified & regulariser families & 0.301 & 0.270 & 0.030 & [0.006, 0.059] \\
    stratified & task lengths & 0.316 & 0.309 & 0.006 & [$-0.008$, 0.021] \\
    stratified & conditions, seeds 2--4 only & 0.179 & 0.174 & 0.006 & [$-0.004$, 0.016] \\
    \bottomrule
  \end{tabular}
\end{table}

Against the additive baseline of the first analysis, the slope lowered the out-of-sample log loss from 0.239 to
0.165 (Table~\ref{tab:prediction}), and the reduction was positive when pooled over held-out seeds, regulariser families
and task lengths, with the training-mode slope, with other fitting windows, and on seeds 2--4 alone, which were not used
for calibration. Most of this gain, however, came from structure the additive baseline could not express. The stratified
baseline, which lets the effects of length and of the early loss differ between task, architecture and initialisation,
reached a log loss of 0.162 without the slope, and adding the slope lowered it only to 0.154, a reduction of 0.008 whose
interval includes zero. Small reductions remained when whole seeds (0.011) or whole regulariser families (0.030) were held
out, but not for held-out lengths or for seeds 2--4. In these models, the early validation loss together with the
condition therefore predicted nearly as well without the slope as with it; the models add one linear slope term and do
not test non-linear or stratum-specific slope effects, and the slope may help when predicting for a regulariser the
model has not seen. Pooled reductions also hide groups in which the slope made predictions worse: with the
stratified baseline, one of the five held-out seeds and two of the six held-out lengths. With reaching the criterion on
the test set as the outcome, the additive-baseline reduction shrank from 0.074 to 0.029 (interval $-0.003$ to 0.057). The slope does not order success consistently across architectures either: the unregularised conventional GRU reached
the criterion on adding at $T = 200$ and 400 with a mean slope of about $-0.26$ after 100 updates, while the
unregularised conventional LSTM did not reach it there within 5000 updates with $-0.13$.

\subsection{Within some conditions, flatter early decay went with the seeds that learned}

\begin{figure}[tbp]
  \centering
  \includegraphics[width=\textwidth]{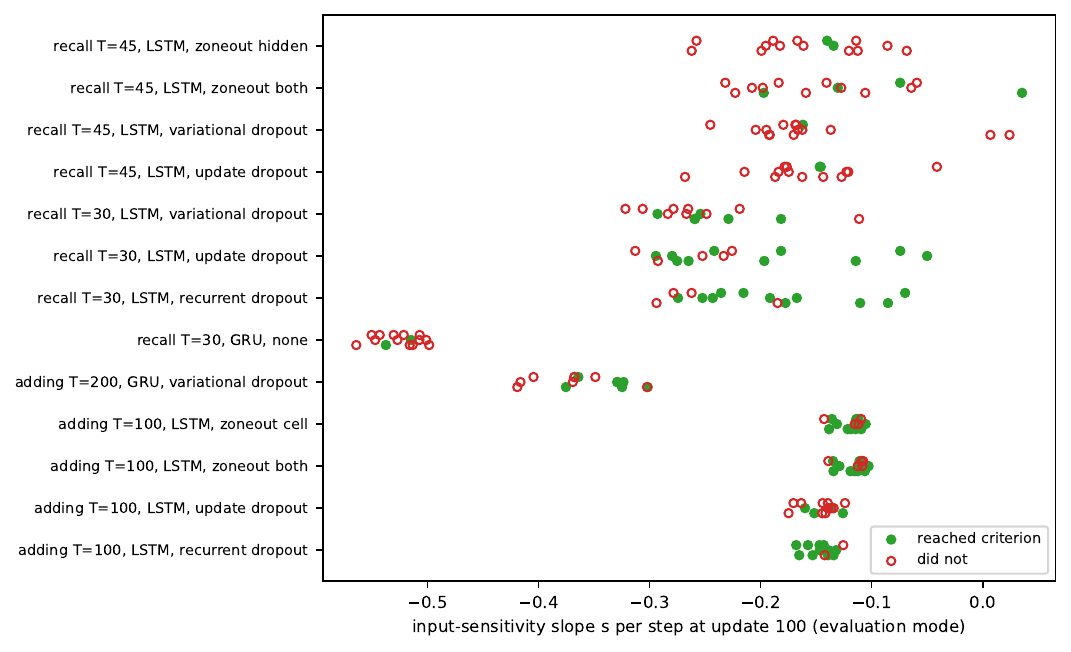}
  \caption{Seeds that learned tended to have flatter early slopes in the LSTM recall conditions, but not in LSTM adding
    at $T = 100$. Confirmatory runs: slope $s$ at update 100 for every run with seeds 5--19 in the 13 conditions with
    mixed outcomes in the main grid. Filled markers reached the criterion within 5000 updates.}
  \label{fig:within}
\end{figure}

Within a condition, runs differ only in their seed, and the question becomes whether the slope tells which seeds will
learn. In the main grid 13 conditions, all conventionally initialised, had both outcomes. Among their 65 runs the slope
ranked runs that reached the criterion above runs that did not with a concordance of 0.735 (95\,\% interval 0.618--0.853
from resampling conditions, 0.556--0.925 from resampling seeds), whereas a lower validation loss at update 100 did not
(0.426, 0.297--0.567). Because this was found after looking at the data, we tested it on 15 new seeds per condition, as
specified beforehand. On these 195 runs (504 pairs of a run that reached the criterion and one that did not) the
concordance was 0.613 for the slope and 0.462 for the early loss (Figure~\ref{fig:within}). With the prespecified interval,
which resamples conditions, the slope passed the test (0.515--0.698) and the loss did not (0.340--0.586). Seeds, however,
are shared across conditions, and resampling them widens the slope interval to 0.482--0.744, and resampling conditions and
seeds crossed to 0.409--0.779; both include 0.5. The paired difference between the two predictors on the same pairs was
0.151 (0.014--0.285 resampling conditions, $-0.015$ to 0.305 resampling seeds). The association was not uniform: the
concordance exceeded 0.6 in all seven conventional LSTM recall conditions and in the GRU adding condition, but was 0.19--0.55
in the four LSTM adding conditions at $T = 100$ and 0.46 for the GRU on recall. The evidence therefore supports a modest,
condition-dependent association rather than a general ranking of seeds; the absence of an association for the early loss
is a failure to detect one, not evidence that none exists. In the hazard model, a slope flatter by one standard deviation
across runs was associated with higher odds of reaching the criterion in each interval (0.93 on the log-odds scale,
interval 0.59--1.54 from refitting on resampled conditions; 720 runs, 20\,954 intervals); because this model uses the
additive covariates, which Section~\ref{sec:prediction} found insufficient across conditions, it is descriptive only.

\section{Discussion}

The clearest result is that gate initialisation produced larger contrasts than any regulariser we tested. The
comparison is between particular levels: a conventional default suited to short dependencies, a horizon-matched chrono
initialisation and one rate of each regulariser, at lengths chosen so that most conventionally initialised conditions
did not reach the criterion at the longest length within 5000 updates. Other rates, tasks or lengths could change the
ranking. With
matched chrono initialisation the input sensitivity attenuated far less with lag than under conventional initialisation,
every run reached the
criterion, and the regulariser changed only how long it took; with conventional initialisation the sensitivity decayed by
orders of magnitude over the sequence and the choice of regulariser could decide whether a run learned within the budget.
For the LSTM on the adding problem this replicates \citet{tallec2018}, and it agrees with mean-field accounts in which
gate biases set the propagation timescales of LSTMs and GRUs \citep{gilboa2019}. That a horizon off by about a factor of
four still worked in 119 of 120 unregularised runs shows that chrono initialisation tolerates this mismatch on tasks with
one long lag; \citet{gu2020} found it sensitive to $T_{\max}$ in other settings, and whether a spread of timescales or a
suitable common bias is what matters cannot be separated without fixed-bias baselines.

Consistent with the gradient-propagation motivation of zoneout \citep{krueger2017}, applying it to the state that
carries memory flattened the input-sensitivity curve. The flattening was similar in size to the direct effect of the
expected mask in evaluation mode before training. Zoneout helped short recall, but its learning benefits were
inconsistent, and it often slowed learning on adding. Dropout generally slowed learning in all three placements and, once training had begun, steepened the
decay, clearly so for the GRU and, in the LSTM, for variational dropout; variational dropout, which keeps one mask per sequence and was derived from approximate variational inference
\citep{gal2016}, steepened it most. At initialisation, in training mode, the same placement flattened the decay, so
measurements before training would have ranked it above the other dropout placements. One plausible reading is that a fixed mask
removes the same recurrent inputs at every step, which the network compensates for by learning faster-decaying
dynamics; we did not test this mechanism.

Across conditions, the validation loss after 100 updates, modelled per architecture, initialisation and task,
predicted learning nearly as well without the gradient slope as with it, so these models give little support for using
gradient probes to predict which conditions will learn. Within conditions, we detected an association between the slope
and the outcome in the pooled analysis and in the prespecified test, but did not detect one for the early
validation loss. The slope
association was modest, occurred only in some conditions, and was not distinguishable from chance after accounting for
the dependence between runs that share a seed. Whether a gradient probe after a few hundred updates can help select the
seeds to continue remains to be tested directly. The within-condition findings concern conventional
initialisation; all chrono-initialised main-grid runs reached the criterion, leaving no outcome variation to predict.

\paragraph{Practical implications.}
The results support chrono initialisation, even with a rough horizon, for one-layer gated networks that must carry
information over many steps. They do not say whether dropout or zoneout improve generalisation, because the design did
not assess that benefit with a fixed, finite training set, and they do not cover non-recurrent dropout on the inputs and outputs. We did not evaluate an
early-stopping policy based on either predictor.

\paragraph{Limitations.}
The study uses one-layer networks with 64 units, one optimiser and learning rate, one rate per regulariser family and
two synthetic tasks with fresh or exhaustive training data, so the study measures the regularisers' effects on
optimisation. The task lengths were chosen on unregularised networks so that most conventionally
initialised conditions do not reach the criterion at the longest length within 5000 updates, which shapes the size of the initialisation contrast. The conventional
GRU, unlike the LSTM, has no bias towards keeping its state, and the zoneout rate of 0.15 is lower for the cell state
and higher for the hidden state than the values \citet{krueger2017} found best on character-level Penn Treebank
(they used 0.15 for both states on permuted MNIST); recall has only eight distinct sequences per length and zero inputs between symbol and query, so
derivatives with respect to those inputs measure sensitivity to perturbations the task never presents. The endpoint is
the first validation success; for 19 of 262 adding runs it was not confirmed on the test set. Seeds 0 and 1 were also
used to choose the lengths, and the stratified baseline was added after a review of the first analysis; the
seeds 2--4 analysis and the confirmatory runs address these points only partly. The measured quantity is the input
sensitivity of projections of the final hidden state, not a direct measurement of the LSTM cell-state Jacobian, and a
single exponential describes chrono curves poorly. The prediction intervals resample conditions and, for the primary
comparison, conditions and seeds, but hold the fitted out-of-sample predictions fixed. The many sensitivity analyses are exploratory; only the within-condition comparison on seeds 5--19
was confirmatory, and its conclusion depends on how the shared seeds are treated.

\section{Conclusion}

We compared gate initialisation with recurrent dropout and zoneout in one-layer LSTM and GRU networks on two synthetic
long-dependency tasks with fresh or exhaustive training data, at one rate per regulariser. Initialisation
produced the largest differences, both in how far the input sensitivity reached and in whether and how fast the networks
learned. Dropout and zoneout mattered mainly when initialisation left the sensitivity decaying fast: dropout generally
slowed learning, and zoneout helped in some settings and not in others. Chrono initialisation kept its advantage with a
horizon about four times too short or too long. The decay measured after 100 updates added little to predicting which
conditions learn once the early validation loss was modelled per task, architecture and initialisation. Within
conditions a flatter early decay was associated with the seeds that went on to learn, but in the test specified
beforehand this association was modest, condition-dependent, and robust only to resampling conditions, not seeds.

\section*{Code and data availability}

The experiment code, the analysis and the per-run results are available at
\url{https://github.com/SyntaxSpirits/rnn-gradient-reach}; the gradient curves of every run are archived at
\url{https://doi.org/10.5281/zenodo.23285386}. Networks were trained with rust-lstm 0.10.0 \citep{rustlstm}.

\bibliography{refs}
\bibliographystyle{tmlr}

\end{document}